\documentclass{article}

\usepackage[preprint]{neurips_2019}

\usepackage[utf8]{inputenc} 
\usepackage[T1]{fontenc}    
\usepackage{hyperref}       
\usepackage{url}            
\usepackage{booktabs}       
\usepackage{amsfonts}       
\usepackage{nicefrac}       
\usepackage{microtype}      

\usepackage{graphicx}
\usepackage[ruled,vlined]{algorithm2e}
\usepackage{booktabs}

\usepackage{natbib}
\setcitestyle{authoryear,open={(},close={)}}
\hypersetup{colorlinks,linkcolor={blue},citecolor={blue},urlcolor={red}}

\title{Semantic Segmentation With Multi Scale Spatial Attention For Self Driving Cars}

\author{%
  Abhinav Sagar\thanks{Website of author - \url{https://abhinavsagar.github.io/}} \\
  Vellore Institute of Technology\\
  Vellore, Tamil Nadu, India\\
  \texttt{abhinavsagar4@gmail.com} \\
  \And
RajKumar Soundrapandiyan \thanks{Website of author - \url{https://sites.google.com/site/rajkumarsrajkumar/}} \\
  Vellore Institute of Technology\\
  Vellore, Tamil Nadu, India\\
  \texttt{rajkumar.s@vit.ac.in} \\
}

\begin{document}

\nocite{*}

\maketitle

\begin{abstract}
In this paper, we present a novel neural network using multi scale feature fusion at various scales for accurate and efficient semantic image segmentation. We used ResNet based feature extractor, dilated convolutional layers in downsampling part, atrous convolutional layers in the upsampling part and used concat operation to merge them. A new attention module is proposed to encode more contextual information and enhance the receptive field of the network. We present an in depth theoretical analysis of our network with training and optimization details. Our network was trained and tested on the Camvid dataset and Cityscapes dataset using mean accuracy per class and Intersection Over Union (IOU) as the evaluation metrics. Our model outperforms previous state of the art methods on semantic segmentation achieving mean IOU value of 74.12 while running at >100 FPS.  
\end{abstract}

\section{Introduction}

Convolutional neural  networks has seen a lot of success in tasks involving classification, detection and segmentation. These include  bounding box object detection, pose estimation, keypoint prediction and image segmentation. CNN-based neural networks advances, such as dropout \citep{srivastava2014dropout} and batch normalization \citep{ioffe2015batch} have helped avoid some of the common challenges faced earlier like the curse of dimensionality and vanishing gradient problem while training neural networks. 

Convolutional networks are now leading many computer vision tasks, including image classification \citep{deng2009imagenet}, object detection \citep{girshick2014rich}, \citep{zhu2015segdeepm} and \citep{liu2015multiclass} and semantic image segmentation \citep{chen2014semantic}, \citep{li2014highly} and \citep{zhao2017pyramid}. Semantic segmentation is also known as scene parsing, which aims to classify each and every pixel present in the image. It is one of the most challenging and important  tasks in computer vision. The famous fully convolutional network (FCN) \citep{long2015fully} for semantic segmentation is based on VGG-Net \citep{simonyan2014very}, which is trained on the famous ImageNet dataset \citep{deng2009imagenet}.

Segmentation task is different from classification task because it requires predicting a class for each and every pixel of the input image, instead of only discrete classes for the whole input images. In order to predict what is present in the image for each and every pixel, segmentation needs to find not only what is in the input image, but also where it is. It has a number of potential applications in
the fields of autonomous driving, video surveillance, medical imaging etc. This is a challenging problem as there is often a tradeoff between accuracy and speed. Since the model eventually needs to be deployed in real world setting, hence both accuracy and speed should be high.

\section{Related Work}

State-of-the-art methods on semantic segmentation have heavily relied on CNN models trained on large labeled datasets. Fully convolutional networks (FCN) trained pixels-to-pixels using skip connections that combines semantic information from a deep, coarse layer with appearance information from a shallow, fine layer to produce accurate and detailed segmentations. Convolution layers with a kernel size of 1$\times$1 take the place of fully connected layers, followed by unpooling layers to recover the spatial resolution of the feature maps. The success of FCN is due to the great improvements in performance and because it showed that CNN can efficiently learn how to make dense class predictions for semantic segmentation.

After FCN, recently proposed models are mainly designed by (1) bringing out novel decoder structure of the networks \citep{girshick2014rich} and \citep{badrinarayanan2017segnet}; (2) adopting more efficient basic classification models \citep{liu2015semantic} and \citep{bittel2015pixel}; (3) adding integrating context knowledge with some independent modules \citep{zhu2015segdeepm} and \citep{ronneberger2015u}. SegNet \citep{badrinarayanan2017segnet} used an alternative decoder variant, in which an encoder decoder convolution path was proposed. Another deconvolution network was used in \citep{noh2015learning} with a similar decoder path as SegNet, but they adopted deconvolution modules to implement upsampling operations. 

\citep{ronneberger2015u} added a 2$\times$2 up-convolution layer, with a concatenation with corresponding pooling layer in U-Net. FCCN \citep{lin2016efficient} could also be regarded as an alternative decoder structure. \citep{chen2018encoder} used atrous spatial pyramid pooling to embed contextual information at various scales which consist of parallel dilated convolutions with different dilation rates. \citep{zhao2017pyramid} used multi-scale contextual information by combining feature maps generated using different dilated convolutions and pooling operations.

\citep{lin2017refinenet} proposed to fuse mid-level and high-level semantic features using an encoder decoder architecture. \citep{paszke2016enet} reduced the number of downsampling times to get an extremely tight fusion structure. \citep{zhao2018icnet} uses multi-scale images as input and a cascade network to raise efficiency. \citep{li2019dfanet}  uses Subnetwork Aggregation and Sub-stage Aggregation to achieve very high FPS and high accuracy using modified Xception bottleneck. \citep{yu2018bisenet} uses spatial path to recover spatial information and to implement real-time calculation.

We summarize our main contributions as follows:

• We propose a new model architecture which used dilated convolutional layers in downsampling part and atrous convolutional layers in upsampling at multiple scales.  

• Concat operator is used for merging the feature maps for context encoding. We also propose our very own attention module which encodes channel wise information to model more contextual information and enlarge the receptive field.

• We present the layer wise details, optimization and ablation study of our neural network.

• On evaluating our network using Camvid dataset and Cityscapes dataset using mean accuracy per class and IOU as evaluation metrics, our model outperforms previous state of the art model architectures while running at > 100 FPS.

\section{Background}

\subsection{Feature Fusion}

Feature fusion have been used successfully in semantic segmentation networks. As the increase of the depth of network, the fusion and reuse of features show significant advantages. The upsampling in decoder recovers the spatial information from the downsampling in encoder. \citep{lin2017refinenet} proposes a refine network module to finely fuse features. \citep{li2019dfanet} proposes two feature fusion methods (Sub-network Aggregation and Substage Aggregation) to enhance feature extraction capabilities. Using feature fusion, the interaction between different layers in terms of spatial information and semantic information improves thus obtaining better results. A depthwise separable convolution layer with kernel size k = 3 is used for feature fusion. 

\subsection{Context Encoding}

\citep{hu2018squeeze} used channel wise attention to take into account channel wise information. This work achieved state of the art results in image classification. The challenge in semantic segmentation is to enhance the receptive field and the classification ability. The goal is to extract more contextual information using multi scale feature maps. We present our own attention based module which handles the aforementioned problem well.

\subsection{Attention Modules}

The advantage of using attention based models are that they are very effective in modelling long range dependencies. \citep{vaswani2017attention} first used self attention module for machine translation. \citep{wang2018non} used attention module for images and video using non local operation. In this work, we use attention module for more efficient image segmentation by taking into account more contextual information thus enlarging the receptive field.

\section{Proposed Method}

\subsection{Dataset}

1. \textbf{Camvid dataset:} The Cambridge-driving Labeled Video Database (CamVid) is a collection of videos with object class semantic labels, complete with metadata. The database provides ground truth labels that associate each pixel with one of 32 classes. The images are of size 360$\times$480. The original images are taken as ground truth. For any algorithm, the metrics are always evaluated in comparison to the ground truth data. The ground truth information is provided in the dataset for the training and test set. A sample image from dataset is shown in Figure 1:

\begin{figure}[htp]
    \centering
    \includegraphics[width=4cm]{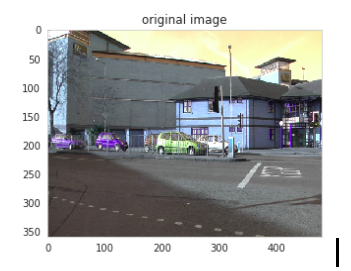}
    \caption{Sample image from dataset.}
    \label{fig3}
\end{figure}

For semantic segmentation problems, the ground truth includes the image, the classes of the objects in it and a segmentation mask for each and every object present in a particular image. Since there is a lot of overlaps in between the labels, hence for the sake of convenience we have gone with 12 labels in this work. These images are shown in binary format for each label separately in Figure
2:

\begin{figure}[htp]
    \centering
    \includegraphics[width=6cm]{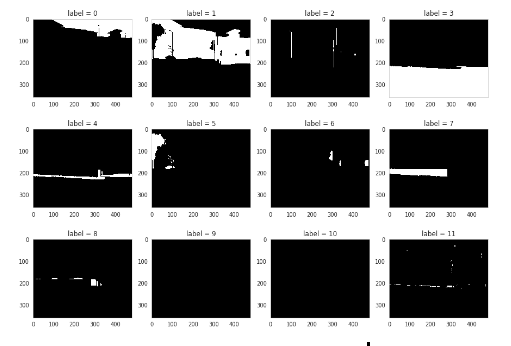}
    \caption{Sample image converted to binary class mask.}
    \label{fig3}
\end{figure}

The classes chosen from the dataset are Sky, Building, Pole, Road, Pavement, Tree, SignSymbol, Fence, Car, Pedestrian and Bicyclist.

2. \textbf{Cityscapes dataset:} This dataset contains urban street scenes images from 50 different cities. The images are divided into 5,000 finely annotated images and 19,998 coarsely annotated images. The total number of classes in the dataset is 30, but we have only used 19 classes for both training and evaluation. The images in training, validation and test set are 2,975, 500, 1,525 respectively.

\subsection{Network Architecture}

We split the dataset into 2 parts with 85 percent images in the training set and 15 percent images in the test set. The loss function used is categorical cross entropy. We used dilated convolutions in downsampling part to reduce the feature maps and atrous convolutions in upsampling part to recover back the features. Concat operation is used to merge the features at different scales thus encoding more contextual information. We use our very own designed attention module for enlarging the receptive field and encode more contextual information. The attention module used in this work is shown in Figure 3:

\begin{figure}[htp]
    \centering
    \includegraphics[width=8cm]{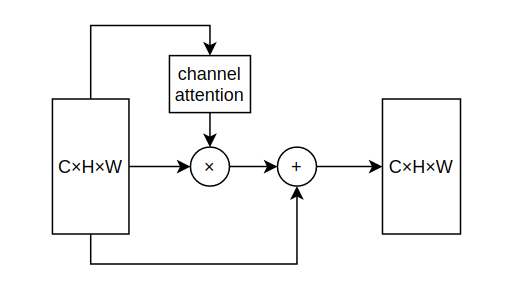}
    \caption{Illustration of our attention module. Here x denotes matrix multiplication and + denotes element wise sum. C, W and H respectively denotes channel, width and height of a layer respectively.}
    \label{fig3}
\end{figure}

For the dilated convolutional layer we didn’t use any padding, used 3$\times$3 filter and use relu as the activation function. For the max pooling layer, we used 2$\times$2 filters and strides of 2$\times$2. ResNet is used as the backbone and the feature extractor. In the upsampling path we used atrous convolutions layers with 4$\times$4 kernel size and strides of 4$\times$4. Softmax is used as the activation function in the last layer to output discrete probabilities of whether an object is present in a particular pixel location or not. We used Adam as the optimizer for training our network. The network architecture used in this work is shown in Figure 4:

\begin{figure}[htp]
    \centering
    \includegraphics[width=10cm]{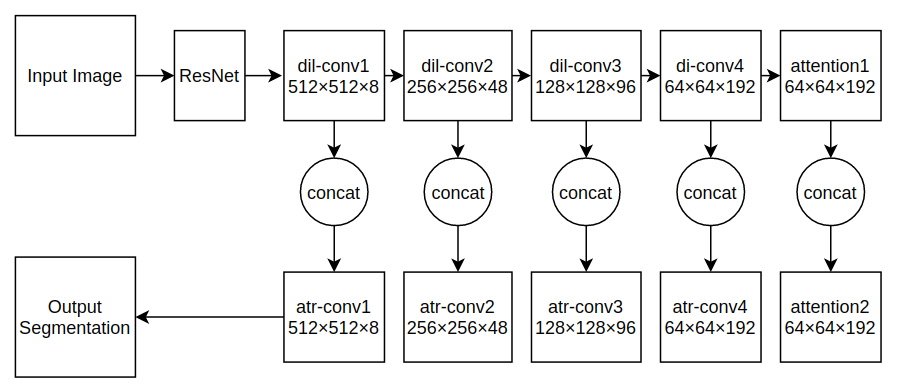}
    \caption{Illustration of our neural network architecture. Here dil-conv represents dilated convolutions and atr-conv represents atrous convolutions. attention1 and attention2 are the two channel wise attention modules used in this work.}
    \label{fig3}
\end{figure}

\subsection{Optimization}

Suppose given a local feature $C$, we feed it into a convolution layers to generate two new feature maps $B$ and $C$ respectively. After that we perform a matrix multiplication between the transpose of $A$ and $B$, and apply a softmax layer to calculate the spatial attention map as shown in Equation 1:

\begin{equation}
s_{j i}=\frac{\exp \left(A_{i} \cdot B_{j}\right)}{\sum_{i=1}^{N} \exp \left(A_{i} \cdot B_{j}\right)}    
\end{equation}

We then perform a matrix multiplication
between the transpose of $X$ and $A$ and reshape their results. Then we multiply the result by a scale parameter $\beta$ and perform an element-wise sum operation with
$A$ to obtain the final output as shown in Equation 2:

\begin{equation}
E_{j}=\alpha \sum_{i=1}^{N}\left(s_{j i} D_{i}\right)+C_{j} 
\end{equation}

The Equation 2 shows that the resultant feature of each channel is a weighted sum of the features of all channels and models the semantic dependencies between feature maps at various scales. For a single backbone $\phi n(x)$, a stage process, the stage in the previous backbone network and sub-stage aggregation method can be formulated as shown in Equation 3: 

\begin{equation}
x_{n}^{i}=\left\{\begin{array}{ll}
x_{n}^{i-1}+\phi_{n}^{i}\left(x_{n}^{i-1}\right) \\
{\left[x_{n}^{i-1}, x_{n-1}^{i}\right]+\phi_{n}^{i}\left(\left[x_{n}^{i-1}, x_{n-1}^{i}\right]\right)}
\end{array}\right.    
\end{equation}

Here $i$ refers to the index of the stage. 

\subsection{Loss Functions}

The loss function used is Softmax loss as shown in Equation 4:

\begin{equation}
Loss =\frac{1}{N} \sum_{i}-\log \left(\frac{e^{p_{i}}}{\sum_{j} e^{p_{j}}}\right)
\end{equation}

where $p$ is the output prediction of the network, i and j are stages in the network.

\subsection{Ablation Studies}

The effect of the number of pooling layers on Intersection Over Union(IOU) using Camvid dataset is shown in Table 1. As can be noted, using more pooling layers increases IOU but it's effect is not consistent.

\begin{table}[h]
  \caption{Results on Camvid dataset with different numbers of pooling in each stage of the backbone, “$\times$N” means the number of pooling.}
  \label{sample-table2}
  \centering
  \begin{tabular}{ll}
  \toprule
    Number of pooling &mIoU(\%)\\
   \midrule
 Pooling $\times$0 &70.4\\
Pooling $\times$1 &71.3\\
Pooling $\times$2 &73.8\\
Pooling $\times$3 &73.4\\
Pooling $\times$4 &74.9\\
Pooling $\times$5 &75.6\\
    \bottomrule
  \end{tabular}
\end{table}

The effect of varying the number of branches and fusion methods used in model architecture on IOU using Cityscapes dataset is shown in Table 2. Using more number of branches and concat fusion instead of not using one increases the IOU. 

\begin{table}[h]
  \caption{Results on Cityscapes dataset with with different number of branches and fusion methods.}
  \label{sample-table3}
  \centering
  \begin{tabular}{lll}
  \toprule
    Number of branches &Fusion methods &mIoU(\%)\\
   \midrule
1 &None &74.4\\
1 &concat &75.8\\
2 &None &75.7\\
2 &concat &77.5\\
    \bottomrule
  \end{tabular}
\end{table}

We investigate the effect of each component in
our proposed network. We use VGG16 and ResNet50 as the base network and evaluate our method on the Camvid validation
dataset as shown in Table 3:

\begin{table}[h]
  \caption{Accuracy and parameter analysis of our baseline model: VGG16 and ResNet50 on Camvid validation dataset. Here we use FCN-32s as the base structure.
FLOPS are estimated for input of size 3$\times$640$\times$360.}
  \label{sample-table7}
  \centering
  \begin{tabular}{lllll}
  \toprule
    Method &BaseModel &FLOPS &Parameters &Mean IOU(\%)\\
   \midrule
FCN-32s &VGG16 &47.5M &1.4M &61.28\\
FCN-32s &ResNet50 &12.5G &22.5M &61.73\\
    \bottomrule
  \end{tabular}
\end{table}

\subsection{Implementation Details}

The initial learning rate was set to 0.001 in all our experiments. Momentum and weight decay coefficients are set to 0.9 and 0.0001 respectively. Batch size of 8 was chosen for all our experiments. Data augmentation operations like shearing, cropping and flipping was performed to artificially increase the dataset size.

\section{Results}

In this section we present the results of our work and compare the  results we achieved with previous state of the art. The model is trained for 40 epochs and reaches a training mean pixel accuracy of 93 percent and validation mean pixel accuracy of 88 percent. The loss and pixel wise accuracy (both training and test) are plotted as a function of epochs in Figure 5:

\begin{figure}[htp]
    \centering
    \includegraphics[width=14cm]{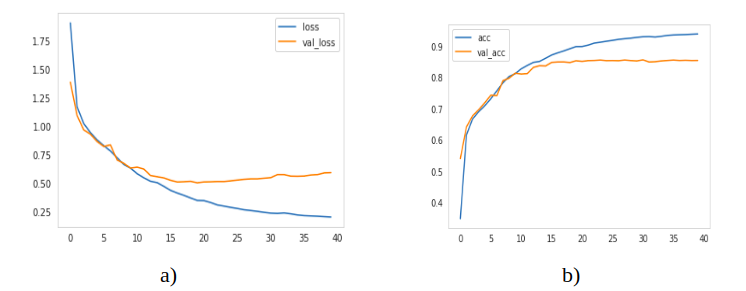}
    \caption{a) Loss vs epochs b) Accuracy vs epochs}
    \label{fig4}
\end{figure}

For evaluating the performance of our model architecture, we used two evaluation metrics:

1. \textbf{Mean Accuracy per-class} - This metric outputs the class wise prediction accuracy per pixel.

2. \textbf{Mean IOU} - It is a segmentation performance parameter that measures the overlap between two objects by calculating the ratio of intersection and union with ground truth masks. This metric is also known as Jaccard Index. 

The class wise IOU values were calculated using Equation 5.

\begin{equation}
I o U=\frac{T P}{(T P+F P+F N)}    
\end{equation}

Where $TP$ denotes true positive, $FP$ denotes false positive, $FN$ denotes false negative and $IOU$ denotes Intersection over union value.

We present the class wise IOU values for all the twelve classes present using CamVid dataset in Table 4.

\begin{table}[h]
  \caption{IOU values for all classes}
  \label{sample-table4}
  \centering
  \begin{tabular}{lllllllllllll}
  \toprule
    Class &1 &2 &3 &4 &5 &6 &7 &8 &9 &10 &11 &12           \\
    \midrule
    IOU &0.923 &0.905 &0.232 &0.947 &0.831 &0.344 &0.569& 0.792 &0.283 &0.261 &0.457 &0.527          \\
    \bottomrule
  \end{tabular}
\end{table}

The effect of using multiple blocks, FLOPS and parameters on IOU using Cityscapes dataset is shown in Table 5. Here FLOPS and parameters are a measure of computation required by our model architecture.

\begin{table}[h]
  \caption{Detailed performance comparison of our proposed aggregation strategy on Cityscapes dataset. ’$\times$N’ means that we replicate N backbones to implement feature aggregation}
  \label{sample-table5}
  \centering
  \begin{tabular}{llll}
  \toprule
    Model &FLOPs(G) &Params(M) &mIoU(\%)\\
   \midrule
    Backbone A &1.4 &2.2 &64.7\\
Backbone A $\times$2 &2.3  &4.5 &65.3\\
Backbone A $\times$3 &2.7 &7.4 &62.1\\
Backbone A $\times$4 &2.9 &10.6 &57.8\\
Backbone B &0.8 &1.4 &59.5\\
Backbone B $\times$2 &1.2 &3.3 &61.5\\
Backbone B $\times$3 &1.4 &4.7 &56.4\\
Backbone B $\times$4 &1.5 &6.1 &52.7\\
    \bottomrule
  \end{tabular}
\end{table}

A comparative analysis using CamVid dataset achieved by previous state of the art model architectures vs ours is shown in Table 6.

\begin{table}[h]
  \caption{Accuracy and speed analysis on CamVid test dataset. Ours is 512$\times$768 input and others are 768$\times$1024 input.}
  \label{sample-table6}
  \centering
  \begin{tabular}{lll}
  \toprule
    Model &Frame(fps) &mIoU(\%)\\
   \midrule
DPN \citep{yu2018learning} &1.2 &60.1\\
DeepLab \citep{chen2017deeplab} &4.9 &61.6\\
ENet \citep{paszke2016enet} &- &51.3\\
ICNet \citep{zhao2018icnet} &27.8 &67.1\\
BiSeNet1 \citep{yu2018bisenet} &- &65.6\\
BiSeNet2 \citep{yu2018bisenet} &- &68.7\\
DFANet A \citep{li2019dfanet} &120 &64.7\\
DFANet B \citep{li2019dfanet} &160 &59.3\\
SwiftNet pyr \citep{orsic2019defense} &- &72.85\\
SwiftNet \citep{orsic2019defense} &- &73.86\\
Ours &124 &74.12\\
    \bottomrule
  \end{tabular}
\end{table}

A comparative analysis using Cityscapes dataset achieved by previous state of the art model architectures vs ours is shown in Table 7.

\begin{table}[h]
  \caption{Accuracy and speed analysis on Cityscapes test dataset.}
  \label{sample-table6}
  \centering
  \begin{tabular}{lllllll}
  \toprule
Model &InputSize &FLOPs &Params &Time(ms) &Frame(fps) &mIoU(\%)\\
   \midrule
PSPNet \citep{zhao2017pyramid} &713 $\times$ 713 &412.2G &250.8M &1288 &0.78 &81.2\\
DeepLab \citep{chen2017deeplab} &512 $\times$ 1024 &457.8G &262.1M &4000 &0.25 &63.1\\
SegNet \citep{badrinarayanan2017segnet} &640 $\times$ 360 &286G &29.5M &16 &16.7 &57\\
ENet \citep{paszke2016enet} &640 $\times$ 360 &3.8G &0.4M &7 &135.4 &57\\
CRF-RNN \citep{zheng2015conditional} &512 $\times$ 1024 &- &- &700 &1.4 &62.5\\
FCN-8S \citep{long2015fully} &512 $\times$ 1024 &136.2G &- &500 &2 &63.1\\
FRRN \citep{pohlen2017full} &512 $\times$ 1024 &235G &- &469 &0.25 &71.8\\
ICNet \citep{zhao2018icnet} &1024 $\times$ 2048 &28.3G &26.5M &33 &30.3 &69.5\\
BiSeNet1 \citep{yu2018bisenet} &768 $\times$ 1536 &14.8G &5.8M &13 &72.3 &68.4\\
BiSeNet2 \citep{yu2018bisenet} &768 $\times$ 1536 &55.3G &49M &21 &45.7 &74.7\\
DFANet A \citep{li2019dfanet} &1024 $\times$ 1024 &3.4G &7.8M &10 &100 &71.3\\
DFANet B \citep{li2019dfanet} &1024 $\times$ 1024 &2.1G &4.8M &8 &120 &67.1\\
Ours &1024 $\times$ 1024 &1.8G &5.5M &6 &134 &72.4\\
    \bottomrule
  \end{tabular}
\end{table}

The results comparing the predicted segmentations and ground truth segmentation using CamVid dataset is shown in Fig 6.

\begin{figure}[htp]
    \centering
    \includegraphics[width=12cm]{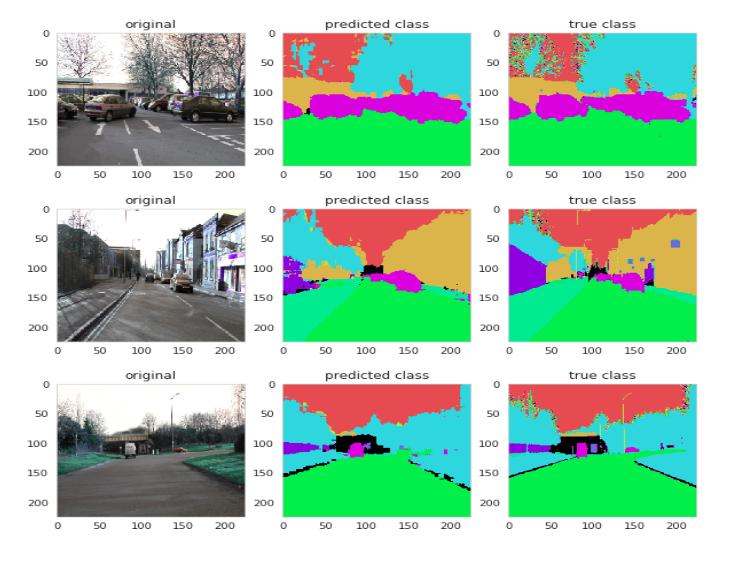}
    \caption{Results using CamVid dataset. First column: input image from dataset, second column: predicted segmentation from our network and third column: ground truth segmentation.}
    \label{fig5}
\end{figure}

\section{Conclusions}

In this paper, we proposed a semantic segmentation network using multi scale attention feature maps and validated its performance on Camvid dataset and Cityscapes dataset. We used a downsampling and upsampling structure with dilated and atrous convolutional layers respectively with combinations between corresponding pooling and unpooling layers. We also propose our own attention module to enlarge the receptive field and encode more contextual information. Multi scale feature maps are merged using concat operator for encoding more contextual information. We present loss function, optimization details, ablation studies and evaluation metrics used. Our network achieves mean IOU value of 74.12 which is better than the previous state of the art on semantic segmentation while running at >100 FPS. 

\subsubsection*{Acknowledgments}

We would like to thank Nvidia for providing the GPUs.

\bibliography{neurips_2019}

\end{document}